\pdfoutput=1

\documentclass[11pt]{article}

\usepackage[preprint]{acl}

\usepackage{times}
\usepackage{latexsym}

\usepackage[T1]{fontenc}

\usepackage[utf8]{inputenc}

\usepackage{microtype}

\usepackage{inconsolata}

\usepackage{graphicx}

\usepackage{inconsolata}
\usepackage{amsmath, amssymb, amsthm}
\usepackage{bm}
\usepackage{makecell}
\usepackage{url}
\usepackage{subfigure}
\usepackage{comment}
\usepackage{booktabs}
\usepackage{makecell}
\usepackage{color}
\usepackage{pifont}
\usepackage{url}
\usepackage{multirow}
\usepackage{tabularx}
\usepackage{colortbl}
\usepackage{lmodern} %
\usepackage{newtxtext} %
\usepackage{listings}       %
\usepackage{array}          %
\usepackage{fancyvrb}
\usepackage{geometry}
\usepackage{tcolorbox}
\usepackage[whole]{bxcjkjatype}
\tcbuselibrary{listings, breakable}
\usepackage{mdframed} %
\usepackage{tablefootnote}
\usepackage{enumitem}
\usepackage{fdsymbol}

\title{llm-jp-modernbert: A ModernBERT Model Trained on a Large-Scale Japanese Corpus with Long Context Length}

\author{
  Issa Sugiura$^{\clubsuit,\diamondsuit}$,
  Kouta Nakayama$^{\diamondsuit}$,
  Yusuke Oda$^{\diamondsuit}$\\
  $^{\clubsuit}$Kyoto University,
  $^{\diamondsuit}$ NII LLMC\\
  \texttt{sugiura.issa.q29@kyoto-u.jp},
  \texttt{\{nakayama, odashi\}@nii.ac.jp}\\
}

\begin{document}
\maketitle
\begin{abstract}

Encoder-only transformer models like BERT are widely adopted as a pre-trained backbone for tasks like sentence classification and retrieval. However, pretraining of encoder models with large-scale corpora and long contexts has been relatively underexplored compared to decoder-only transformers. In this work, we present llm-jp-modernbert, a ModernBERT model trained on a publicly available, massive Japanese corpus with a context length of 8192 tokens. 
While our model does not surpass existing baselines on downstream tasks, it achieves good results on fill-mask test evaluations. We also analyze the effect of context length expansion through pseudo-perplexity experiments. Furthermore, we investigate sentence embeddings in detail, analyzing their transitions during training and comparing them with those from other existing models, confirming similar trends with models sharing the same architecture. To support reproducibility and foster the development of long-context BERT, we release our model\footnote{\url{https://huggingface.co/llm-jp/llm-jp-modernbert-base}}, along with the training and evaluation code\footnote{\url{https://github.com/llm-jp/llm-jp-modernbert}}.

\end{abstract}

\section{Introduction}
Encoder-only transformer models, such as BERT~\cite{devlin-etal-2019-bert}, are pre-trained on a large corpus using a masked language modeling (MLM) objective. They are commonly used as pre-trained backbones for a variety of downstream tasks, including sentence classification~\cite{penedo2024the} and sentence retrieval~\cite{lewis2020retrieval}. Since the release of BERT~\cite{devlin-etal-2019-bert}, there have been numerous advancements in model architecture, training methods, and context length~\cite{liu2019robertarobustlyoptimizedbert,he2021deberta,warner2024smarterbetterfasterlonger,breton2025neobertnextgenerationbert}. In parallel, considerable efforts have been made to develop Japanese BERT models~\cite{tohoku2023,waseda2022,ku2024}. Recent efforts have led to the development of modernbert-ja-130m~\cite{modernbert-ja}, a ModernBERT~\cite{warner2024smarterbetterfasterlonger} model trained on in-house Japanese and English corpora with a context length of 8192 tokens.

On the other hand, research on pretraining encoder-only transformer models with large-scale corpora and long contexts has been less active compared to that of decoder-only transformer models. 
This limits our understanding of model behavior during training in such settings.
In addition, few existing models publicly release all components such as training code, evaluation code, and training data, which makes detailed analysis challenging.

In this paper, we introduce llm-jp-modernbert, a ModernBERT model trained on a publicly available, massive Japanese corpus with a context length of 8192. 
To deepen our understanding of model behavior, we analyze training checkpoints with a focus on three aspects: performance on downstream tasks, the effects of context length expansion, and the evolution of sentence embeddings obtained via mean pooling.
By releasing our training code, evaluation code, and model, we aim to foster future research in this area.

\section{Training}
\subsection{Model Architecture} 
The architecture of llm-jp-modernbert is based on ModernBERT-base~\cite{warner2024smarterbetterfasterlonger}, a model that integrates several recent advancements commonly used in large language models (LLMs), such as Rotary Positional Embedding (RoPE)~\cite{su2023roformerenhancedtransformerrotary}, Local-Global Alternating Attention~\cite{team2024gemma}, and FlashAttention~\cite{dao2022flashattention}.

For tokenization, we use a modified version of the llm-jp-tokenizer v3\footnote{\url{https://github.com/llm-jp/llm-jp-tokenizer}}, customized for the encoder model. 
This tokenizer is trained on data from the domains of Japanese, English, Code, Chinese, and Korean, and has a vocabulary size of 99,574.
Consequently, the embedding layer has a larger number of parameters than typical models, resulting in a total of 187M parameters for the model.

\subsection{Training Data}

For the training dataset, we use the Japanese subset of the llm-jp-corpus v4\footnote{The llm-jp-corpus v4 will be publicly available soon.}, which contains approximately 0.69T tokens, tokenized using llm-jp-tokenizer v3.
The llm-jp-corpus v4 was developed by~\citet{llmjp2024llmjpcrossorganizationalprojectresearch} and includes data crawled from sources such as Common Crawl\footnote{\url{https://commoncrawl.org}}, WARP\footnote{\url{https://warp.ndl.go.jp}}, Wikipedia, KAKEN\footnote{\url{https://kaken.nii.ac.jp}}, patents, legal documents, and National Diet proceedings, and more.

\subsection{Training Settings}

We employ a two-stage pretraining approach: In the first stage, the model is pretrained with a maximum context length of 1024 tokens. In the second stage, the context length is extended to 8192 tokens.
Table~\ref{table:training_setting} summarizes the hyperparameters for each stage, which were selected based on and RoBERTa~\cite{liu2019robertarobustlyoptimizedbert}. 
Following~\citet{warner2024smarterbetterfasterlonger}, we set the mask rate to 30\% for the Masked Language Modeling (MLM) objective and omit the Next-Sentence Prediction objective.

The model consumes up to 1.7T tokens during Stage 1, including padding tokens (500k steps $\times$ 3328 $\times$ 1024). The same applies to Stage 2, consuming 0.6T tokens.

\begin{table}[t]
\centering
\begin{tabular}{l|cc}
\toprule
\textbf{Hyperparameters} & \textbf{Stage 1} & \textbf{Stage 2}\\
\hline
Max sequence length&1024 & 8192\\
Training steps & 500,000 & 200,000 \\
Total batch size & 3328 & 384 \\
Peak learning rate& $5 \times 10^{-4}$ & $5 \times 10^{-5}$ \\
Warmup steps & \multicolumn{2}{c}{24,000}\\
LR schedule & \multicolumn{2}{c}{Linear decay} \\
Optimizer & \multicolumn{2}{c}{AdamW} \\
Adam $\beta_1$ & \multicolumn{2}{c}{0.9}\\
Adam $\beta_2$ & \multicolumn{2}{c}{0.98} \\
Adam $\epsilon$ & \multicolumn{2}{c}{$1\times 10^{-6}$} \\
MLM probability & \multicolumn{2}{c}{0.30} \\
Gradient clip & \multicolumn{2}{c}{1.0} \\
Weight decay & \multicolumn{2}{c}{$1\times 10^{-5}$} \\
Global RoPE theta & \multicolumn{2}{c}{10,000} \\
Line by line & \multicolumn{2}{c}{True} \\
Training time & 8 days& 3 days\\
\bottomrule
\end{tabular}
\caption{Training settings. Line by line indicates whether to discard the part exceeding maximum sequence length. We used 16 NVIDIA H100 80GB GPUs for each stage.}
\label{table:training_setting}
\end{table}

\subsection{Training Script}
We prepared a training script based on Hugging Face’s example code\footnote{\url{https://github.com/huggingface/transformers/blob/main/examples/pytorch/language-modeling/run_mlm_no_trainer.py}}, modifying it to support checkpoint resumption with \texttt{IterableDataset}\footnote{\url{https://huggingface.co/docs/datasets/main/en/package_reference/main_classes\#datasets.IterableDataset}}, which we use to handle terabyte-scale datasets.

\begin{figure*}[t]
\small
    \centering
    \begin{minipage}{0.48\textwidth}
        \centering
        \includegraphics[width=\linewidth]{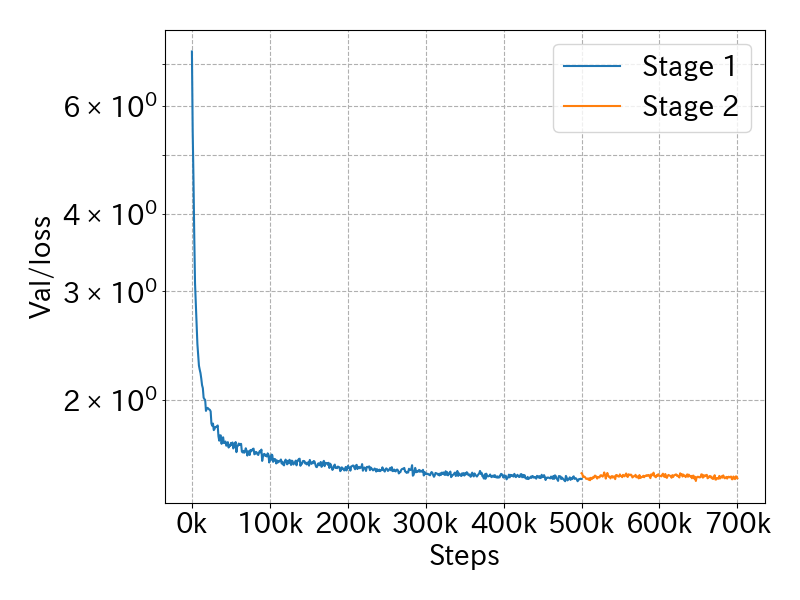}
        \caption*{Loss on validation data}
    \end{minipage}
    \hfill
    \begin{minipage}{0.48\textwidth}
        \centering
        \includegraphics[width=\linewidth]{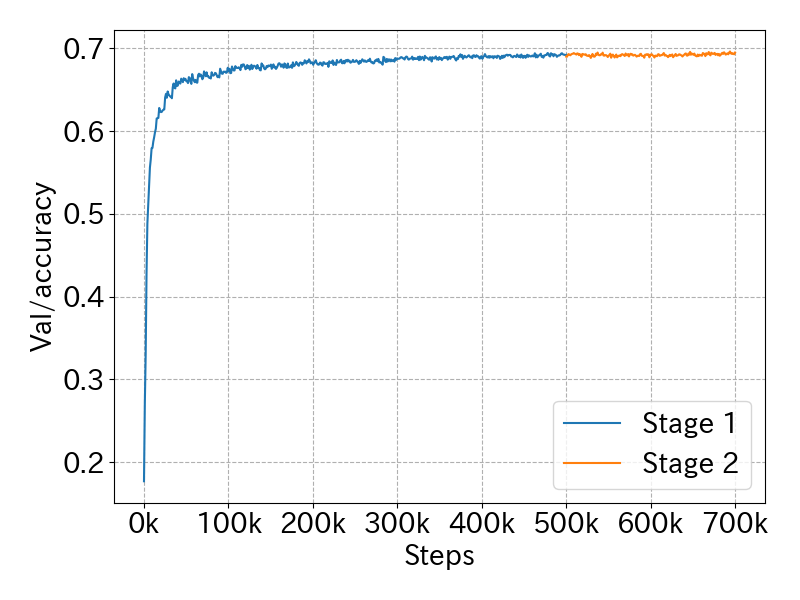}
        \caption*{Accuracy on validation data}
    \end{minipage}
    \vspace{1em}
    \begin{minipage}{0.48\textwidth}
        \centering
        \includegraphics[width=\linewidth]{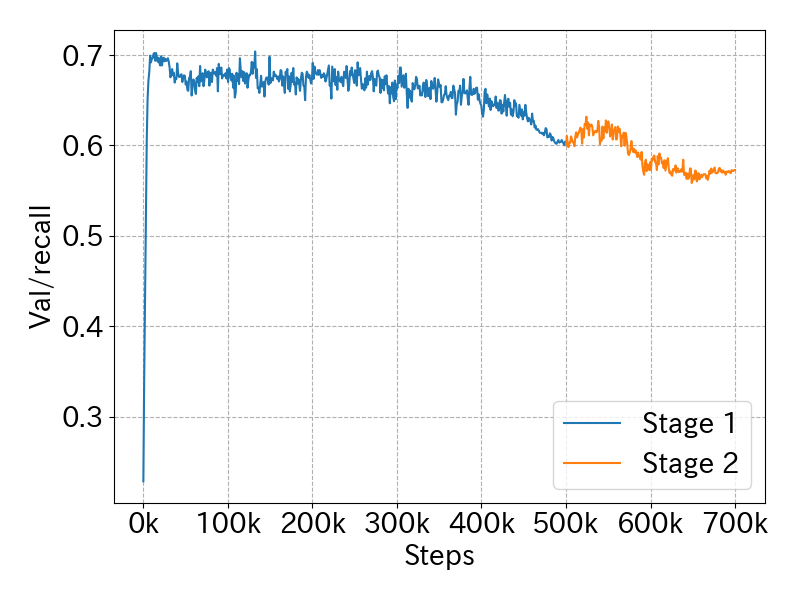}
        \caption*{Recall@10 on MIRACL}
    \end{minipage}
    \hfill
    \begin{minipage}{0.48\textwidth}
        \centering
        \includegraphics[width=\linewidth]{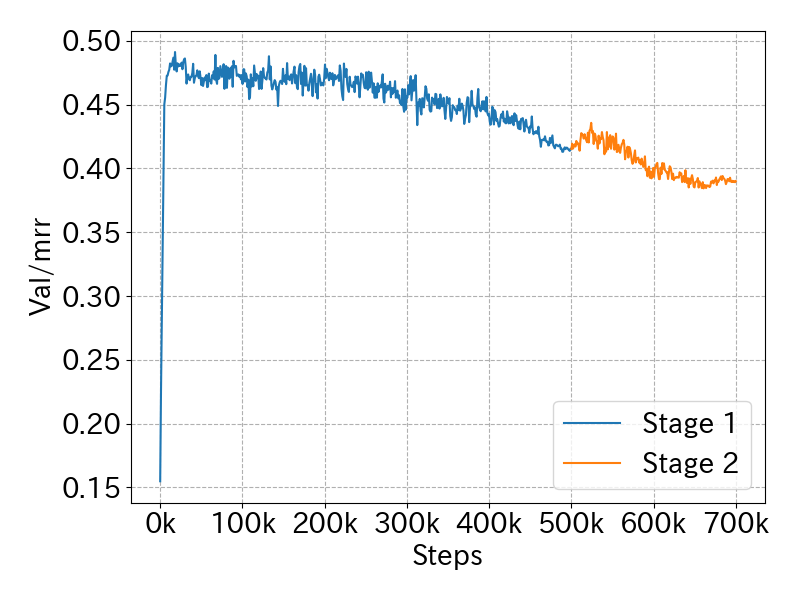}
        \caption*{MRR@10 on MIRACL}
    \end{minipage}
    \caption{Training curves.}
    \label{fig:training_dynamics}
\end{figure*}

\begin{table*}[t]
\centering
\renewcommand{\arraystretch}{1.3}
\small
\begin{tabular}{lcccccccc}
\toprule
\multicolumn{3}{c}{\textbf{Model}} & \textbf{\# Params} &\textbf{JSTS} & \textbf{JNLI} & \textbf{JCoLA} & \textbf{Avg.} \\
\midrule
\multicolumn{3}{l}{tohoku-nlp/bert-base-japanese-v3~\cite{tohoku2023}} & 111M & 92.0 & 91.2 & 88.0 & 90.4 \\
\multicolumn{3}{l}{sbintuitions/modernbert-ja-130m~\cite{modernbert-ja}} & 132M &91.6 & 92.7 & 86.8 & 90.4 \\
\multicolumn{3}{l}{sbintuitions/modernbert-ja-310m~\cite{modernbert-ja}} & 315M & \textbf{93.2} & \textbf{93.3} & \textbf{88.3} & \textbf{91.6} \\
\hline
 & \textbf{Stage} & \textbf{Steps}  & & && & \\
\cline{2-8} 
\multirow{9}{*}{llm-jp-modernbert-base} & \multirow{9}{*}{1} & 4k   &\multirow{9}{*}{187M} & 77.7 & 68.4 & 83.9 & 76.7 \\
               &  & 15k  & & 90.5 & 89.0 & 84.3 & 87.9 \\
                & & 50k  & & 92.1 & 92.0 & 86.2 & 90.1 \\
                  &  & 100k & & 92.1 & 91.8 & 86.1 & 90.0 \\
                   &  & 200k & & 92.0 & 92.7 & 85.0 & 89.9 \\
                    &  & 300k & & 92.0 & 91.9 & 85.2 & 89.7 \\
                     & & 400k & & 92.1 & 92.0 & 85.5 & 89.9 \\
                    & & 500k & & 92.1 & 92.0 & 84.5 & 89.5 \\
                  & 2 & 200k & & 91.8 & 91.3 & 84.4 & 89.2 \\
\bottomrule
\end{tabular}
\caption{Downstream performance on subtasks of JGLUE~\cite{kurihara-etal-2022-jglue}.}
\label{table:jglue}
\end{table*}

\begin{table*}[t]
\centering
\small
\tabcolsep 3pt
\begin{tabular}{p{8cm}|p{2cm}p{2cm}p{2cm}}
\hline
\textbf{Question} & \textbf{bert-base-japanese-v3} & \textbf{modernbert-ja-130m} & \textbf{llm-jp-modernbert-base}\\
\hline
\{ \} は、地球上で最も高い山として知られ、世界中の登山家たちの憧れの地となっています。(\{ \} is known as the highest mountain on earth and is the dream destination for mountaineers from all over the world.)& 現在 (Present) & 富士山 (Mt. Fuji) & エベレスト(Mt. Everest) \\
\hline
\{ \}は、歴史上の重要な出来事であり、多くの人々の生活や社会のあり方に大きな影響を与えました。(\{ \} was an important historical event that had a profound impact on the lives of many people and on the state of society.) & これ (This) & 明治維新 (Meiji Restoration) &  COVID-19 \\
\hline
最も長い川は\{ \}です。その流域には多くの都市や村が広がり、豊かな自然や文化が育まれています。(The longest river is the \{ \}. Many cities and villages are spread out along its basin, nurturing a rich natural environment and culture.)& 川 (River) & 利根川 (Tone River)  &  長江 (Yangtze River) \\
\hline
1年は\{ \}ヶ月です。(One year is \{ \} months.) & 3 & 6 & 12 \\
\hline
英語で「ありがとう」は\{ \}と言います。(In English, 「thank you」 is said as \{ \}.)& ありがとう (Thank you) & サンキュー (Thank you) & サンキュー (Thank you) \\
\hline
1年はおよそ\{ \}日です。(One year is approximately \{ \} days.)& 100 & 7 & 365 \\
\hline
\end{tabular}
\caption{Fill-mask test. \{\} represents the masked token. We filled the mask with the model's top 1 prediction. For llm-jp-modernbert-base, we used the checkpoint after Stage 2 training.}
\label{tab:cloze_test}
\end{table*}

\section{Evaluation}
We evaluate our model from various aspects, including downstream tasks, the impact of context length expansion, and the evolution of sentence embeddings obtained through mean pooling. 
\subsection{Baseline Models}
In this evaluation, we use the following baseline models:

\begin{itemize}
    \item \textbf{tohoku-nlp/bert-base-japanese-v3~\cite{tohoku2023}}: A Japanese BERT model trained on the Japanese portion of the CC-100~\cite{conneau-etal-2020-unsupervised} dataset and the Japanese version of Wikipedia, with a maximum context length of 512.
    \item \textbf{sbintuitions/modernbert-ja-\{130m, 310m\}~\cite{modernbert-ja}}: Japanese ModernBERT models trained on an in-house large-scale corpus of both Japanese and English text, with a maximum context length of 8192.
    \item \textbf{cl-nagoya/ruri-large-v2~\cite{tsukagoshi2024Ruri}}: A supervised fine-tuned Japanese sentence embedding model. This model is used in our experiments related to sentence embeddings.
\end{itemize}

\subsection{Training Curve}
During training, we track multiple validation metrics, including masked language modeling (MLM) loss and accuracy on a validation dataset, as well as recall and Mean Reciprocal Rank (MRR) for a zero-shot sentence retrieval task\footnote{Note that the performance of a supervised fine-tuned model for sentence embedding tasks does not necessarily correlate with that of the pretrained model~\cite{reimers-gurevych-2019-sentence, gao-etal-2021-simcse,fuster-baggetto-fresno-2022-anisotropy}.}.

For MLM loss and accuracy, we use the Japanese validation subset of the llm-jp-corpus-v3, which includes a portion of Wikipedia as the validation dataset. For the zero-shot sentence retrieval task, we use the MIRACL dataset~\cite{zhang2023miracl}, a benchmark for multilingual sentence retrieval\footnote{Details on the construction and validation of the task are provided in Appendix \ref{sec:appendix_miracl}.}. To obtain sentence embeddings during training, we apply mean pooling, which averages the embeddings of all tokens in the sentence to produce a single vector representation for the entire sentence\footnote{We use SentenceBERT~\cite{reimers-2019-sentence-bert}.}. If the input sentence exceeds the maximum sequence length, it is truncated accordingly.

Figure~\ref{fig:training_dynamics} illustrates the training dynamics across different stages. In Stage 1, validation loss and accuracy steadily improve as training progresses. In Stage 2, both metrics show minor improvements, though the increased maximum token length in Stage 2 makes direct loss comparisons with Stage 1 less straightforward. For the sentence retrieval task, performance sharply improves up to 15k steps in Stage 1, after which it gradually declines.

\begin{figure*}[t]
    \centering
    \begin{minipage}{0.32\textwidth}
        \centering
        \includegraphics[width=\linewidth]{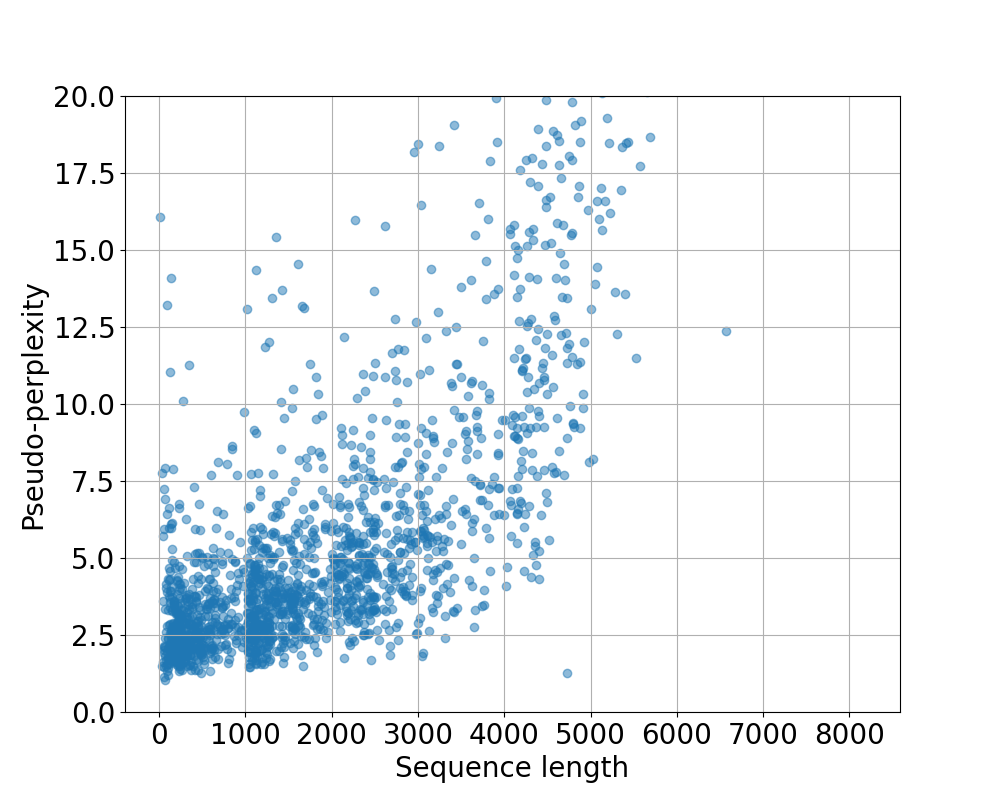}
        \caption*{500k steps in Stage 1}
        \label{fig:pseudo_stage1}
    \end{minipage}
    \hfill
    \begin{minipage}{0.32\textwidth}
        \centering
        \includegraphics[width=\linewidth]{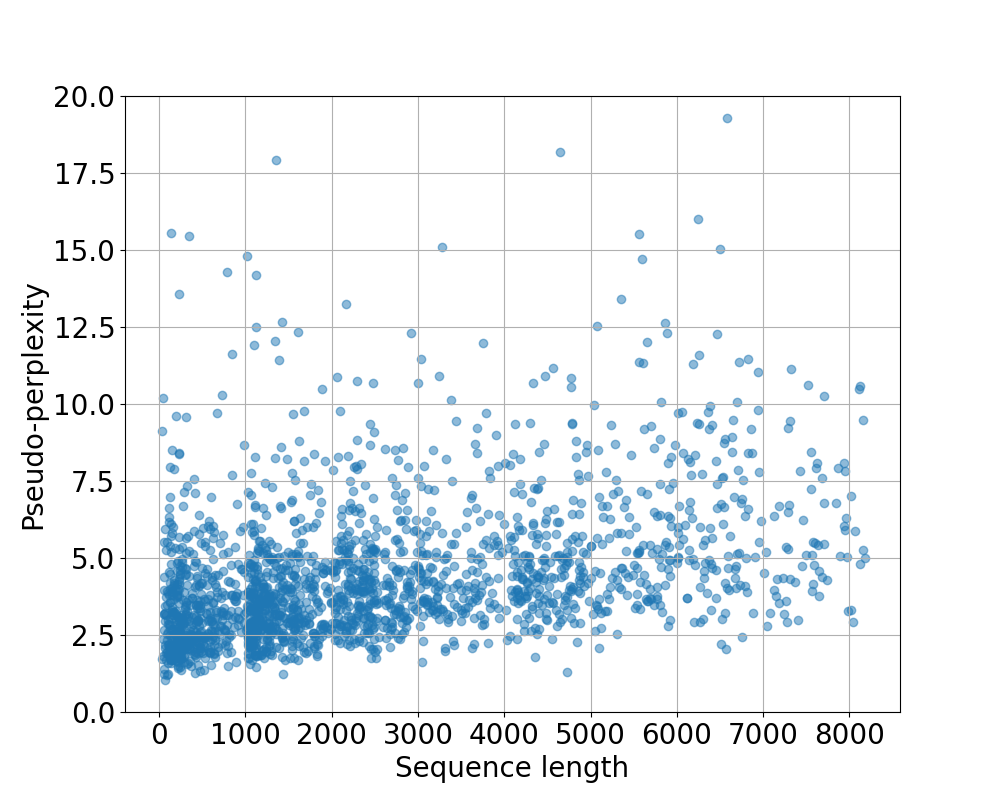}
        \caption*{200k steps in Stage 2}
        \label{fig:pseudo_stage2}
    \end{minipage}
    \hfill
    \begin{minipage}{0.32\textwidth}
        \centering
        \includegraphics[width=\linewidth]{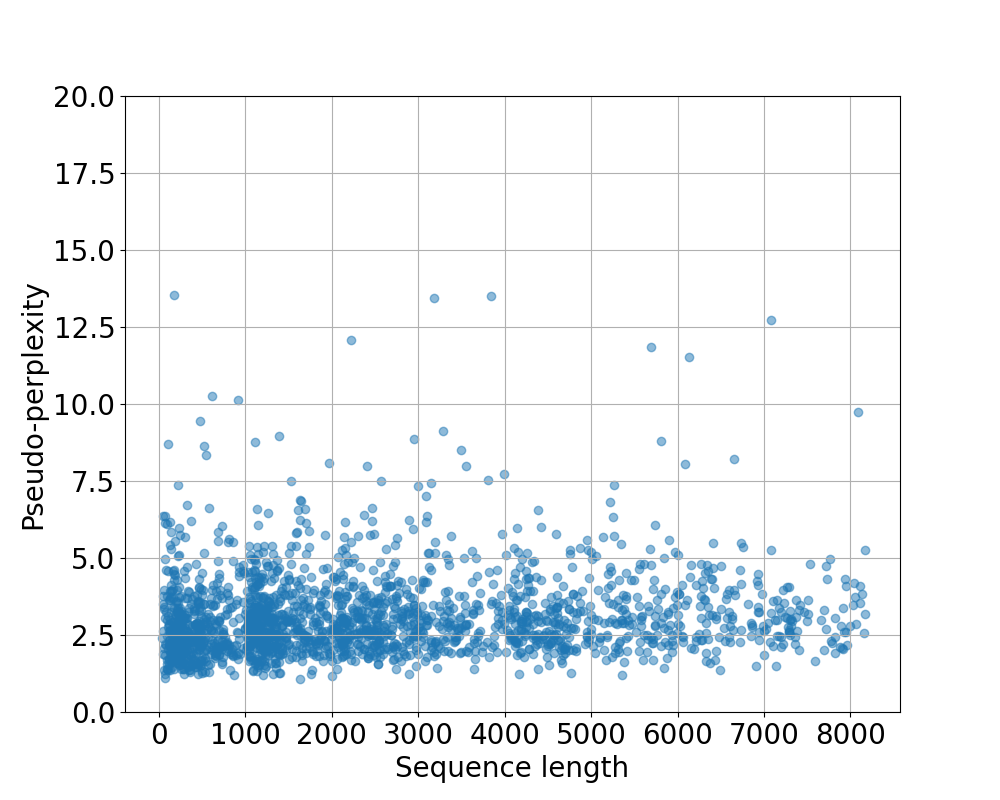}
        \caption*{sbintuitions/modernbert-ja-130m}
        \label{fig:pseudo_sbint}
    \end{minipage}
    \caption{Pseudo-Perplexity vs. Sequence Length.}
    \label{fig:pseudo-ppl}
\end{figure*}

\subsection{Downstream Evaluation}
BERT models are typically pre-trained and then fine-tuned for downstream tasks~\cite{devlin-etal-2019-bert}.
To evaluate the downstream performance of our pre-trained model, we fine-tune and evaluate it on the following tasks from JGLUE~\cite{kurihara-etal-2022-jglue}.

\paragraph{Sentence Classification Task}
For the sentence classification task, we use JCoLA. JCoLA (Japanese Corpus of Linguistic Acceptability) is a binary classification task that determines whether a given sentence is linguistically acceptable.

\paragraph{Sentence Pair Classification Tasks}
For the sentence pair classification task, we use JSTS and JNLI. JSTS (Japanese Semantic Textual Similarity) predicts the semantic similarity between two sentences, while JNLI (Japanese Natural Language Inference) predicts the inference relationship between a premise and a hypothesis sentence. The possible relationships are entailment, contradiction, or neutral.

We use a modified version of Hugging Face's GLUE~\cite{wang2018glue} evaluation code\footnote{\url{https://github.com/huggingface/transformers/blob/main/examples/pytorch/text-classification/run_glue_no_trainer.py}} to support JGLUE. The train split is used for fine-tuning, and the validation split for evaluation. We report the best scores across all combinations of learning rates $\{5 \times 10^{-6}, 1 \times 10^{-5}, 2 \times 10^{-5}, 3 \times 10^{-5}, 5 \times 10^{-5}, 1 \times 10^{-4}\}$ and epochs \{1, 2, 3, 4, 5, 10\}.

As shown in Table~\ref{table:jglue}, JSTS performed well of similar sizes, whereas JCoLA had lower performance.
In Stage 1, JGLUE performance showed no improvement beyond step 50k.

During training, validation loss and accuracy consistently improved, but JGLUE performance plateaued after 50k steps.
Investigating the cause of this discrepancy remains a future challenge.

\subsection{Fill-Mask Test}

The fill-mask test is a task where words in a sentence are masked, and the model is required to predict the masked words. This task is tokenizer-dependent, but It is useful for directly measuring the model's performance on the MLM task. In this experiment, we evaluate BERT models using fill-mask tests on various sentences.
Table~\ref{tab:cloze_test} shows the result. Our model appears to predict the correct words in many examples. Since llm-jp-modernbert is trained on llm-jp-corpus v4, which contains the latest corpus, it is capable of recognizing recent events such as COVID-19.

\subsection{Effect of Context Length Expansion}
\label{sec:pseudo_ppl}
JGLUE mainly consists of short sentences, making it unsuitable for evaluating long-context performance.
Therefore, we conduct a pseudo-perplexity experiment following the methodology introduced in NeoBERT~\cite{breton2025neobertnextgenerationbert}. We sample 2,000 sequences of varying lengths from the Japanese subset of Wikipedia\footnote{We used the train split of the 20231101.ja subset from \url{https://huggingface.co/datasets/wikimedia/wikipedia}.}, stratified into four length-based bins: [0, 1024], (1024, 2048], (2048, 4096], and (4096, 8192) tokens, with 500 sequences selected per bin\footnote{The distribution of sequence lengths for the sampled sequences is provided in Appendix~\ref{sec:appendix_pseudo_ppl}.}. For each sequence, we compute pseudo-perplexity by randomly sampling 100 token positions with replacement, computing the masked language modeling (MLM) loss at each position, and averaging the results. The pseudo-perplexity is defined as 
$P = \exp \left( \frac{1}{n} \sum_{i=1}^{n} l_i \right)$, where $l_i$ is the cross-entropy loss at position $i$ and $n$ is the number of tokens. 

Figure~\ref{fig:pseudo-ppl} presents the results for each model. Consistent with the findings of~\citet{breton2025neobertnextgenerationbert}, 
the pseudo-perplexity for long sequences decreases from Stage 1 to Stage 2, indicating improved performance on extended contexts as a result of the context length expansion introduced in Stage 2. However, our model at 200k steps in Stage 2 shows a slight increase in pseudo-perplexity as sequence length grows, whereas the modernbert-ja-130m maintains consistently low values.
These observations suggest a potential undertraining of our model on long sequences. 
One possible contributing factor might be that Stage 2 training did not explicitly account for the distribution of sentence lengths in the dataset.

\subsection{Alignment and Uniformity}
BERT models are often extended into sentence embedding models through supervised fine-tuning, and alignment and uniformity are commonly used to evaluate how well the model represents sentences~\cite{gao-etal-2021-simcse}. 
While alignment and uniformity do not correlate between pretrained and supervised fine-tuned models, observing sentence embedding behavior during training provides a good way to assess representation quality. Therefore, in this experiment, we measure alignment and uniformity during training.

\paragraph{Alignment}
Alignment is a metric that quantifies how well semantically related positive pairs are positioned close to each other in the embedding space. It is defined as:
\begin{equation}
\ell_{\text{align}} \triangleq \underset{(x, x^+) \sim p_{\text{pos}}}{\mathbb{E}} \| f(x) - f(x^+) \|^2,
\end{equation}
where $p_{\text{pos}}$ is the distribution of positive pairs, and $f$ is a function that embeds text into a normalized vector space.
A smaller value of $\ell_{\text{align}}$ indicates that positive pairs are embedded closer together in the embedding space.

\paragraph{Uniformity}
Uniformity is a metric that measures how evenly sentence embedding vectors are distributed across the embedding space. It is defined as:
\begin{equation}
\ell_{\text{uniform}} \triangleq \log \underset{(x, y) \overset{\text{i.i.d.}}{\sim} p_{\text{data}}}{\mathbb{E}}  e^{-2 \| f(x) - f(y) \|^2}, 
\end{equation}
where $p_{\text{data}}$ is the data distribution. A smaller value indicates that the embeddings are more evenly distributed across the space, reducing bias and preventing excessive concentration in specific regions.

Alignment and uniformity exhibit a trade-off relationship. In an extreme case where all sentences are mapped to the same point, alignment reaches its minimum value of zero, while uniformity attains its maximum value of zero. Conversely, if embeddings are randomly scattered in different directions, uniformity decreases while alignment increases.

To compute $\ell_{\text{align}}$ and $\ell_{\text{uniform}}$, we construct positive pairs and randomly sample sentence pairs (hereafter referred to as random pairs). Specifically, we extract 1746 positive pairs from the MIRACL dataset and sample 2000 random pairs from the Japanese subset of Wikipedia\footnote{We used the train split of the 20231101.ja subset from \url{https://huggingface.co/datasets/wikimedia/wikipedia}}.
The positive pairs are used to compute $\ell_{\text{align}}$, while the random pairs are used to compute $\ell_{\text{uniform}}$.

\begin{figure}[t]
    \centering
    \includegraphics[width=\linewidth]{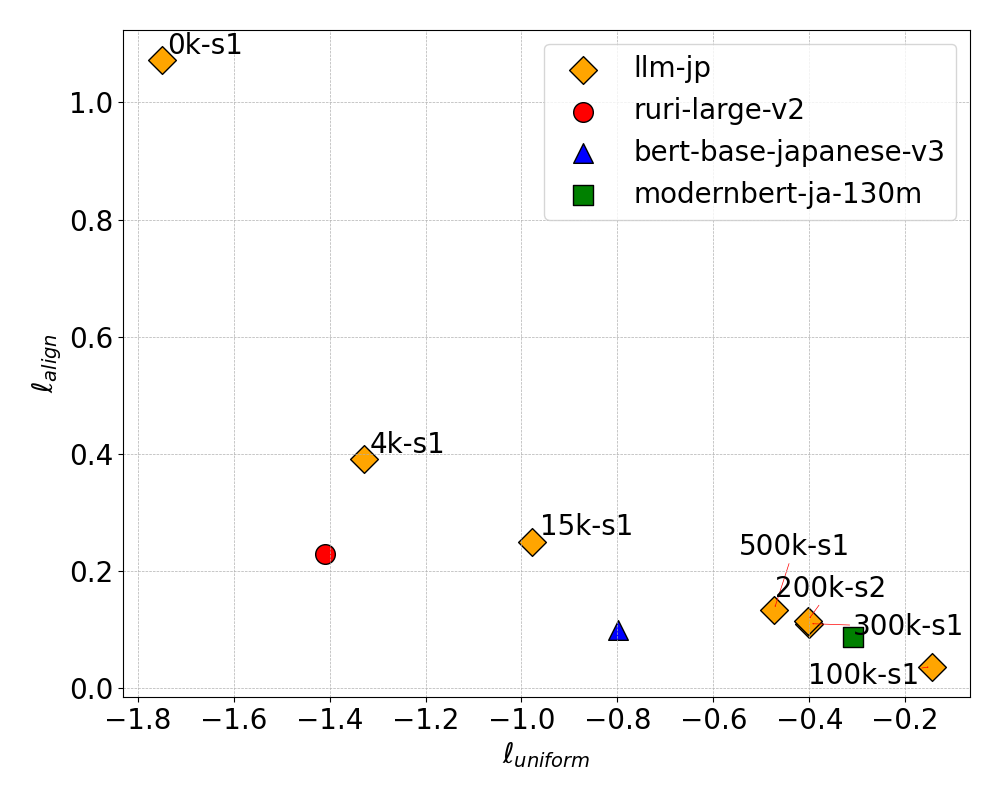}
    \caption{Alignment and uniformity. s1 and s2 represent Stage 1 and Stage 2, respectively.}
    \label{fig:alignment_and_uniformity}
\end{figure}

\begin{figure*}[t]
    \centering
    \begin{minipage}{0.32\textwidth}
        \centering
        \includegraphics[width=\linewidth]{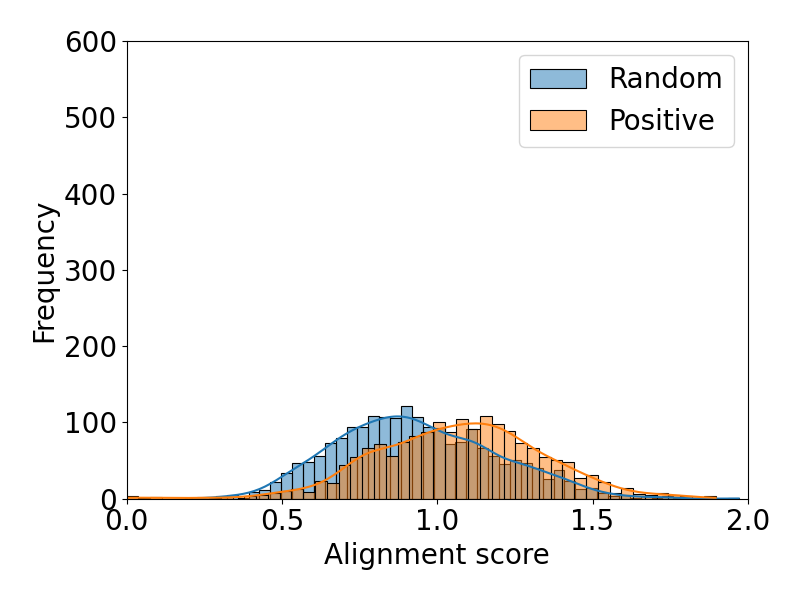}
        \caption*{0k steps in Stage 1}
    \end{minipage}
    \hfill
    \begin{minipage}{0.32\textwidth}
        \centering
        \includegraphics[width=\linewidth]{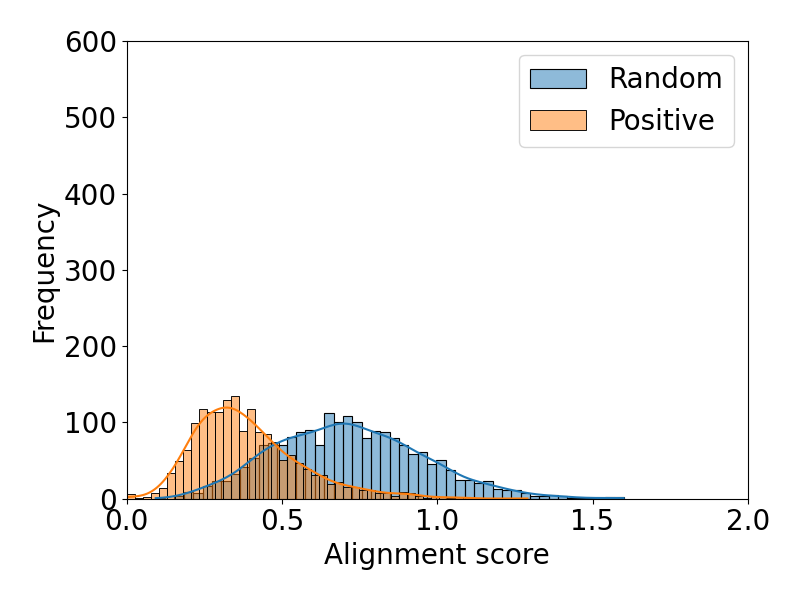}
        \caption*{4k steps in Stage 1}
    \end{minipage} 
    \hfill
    \begin{minipage}{0.32\textwidth}
        \centering
        \includegraphics[width=\linewidth]{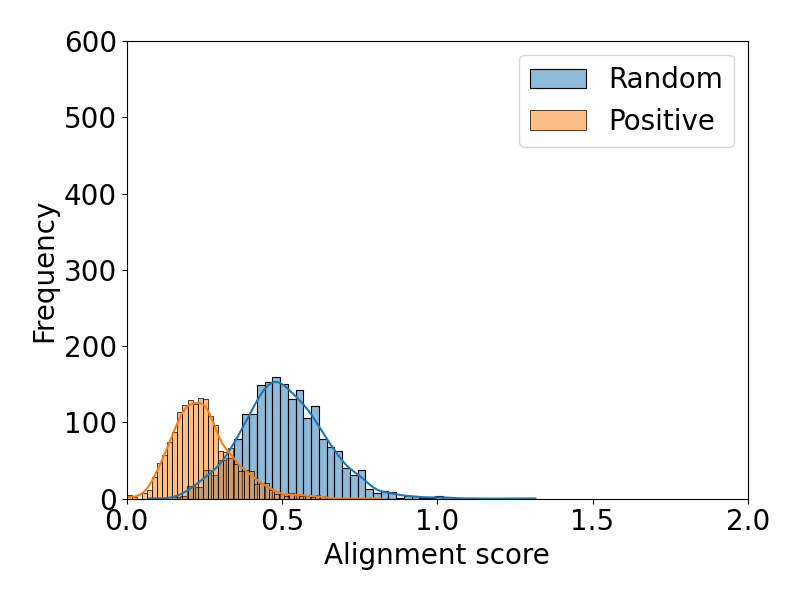}
        \caption*{15k steps in Stage 1}
    \end{minipage}
    \vspace{1em}
    \begin{minipage}{0.32\textwidth}
        \centering
        \includegraphics[width=\linewidth]{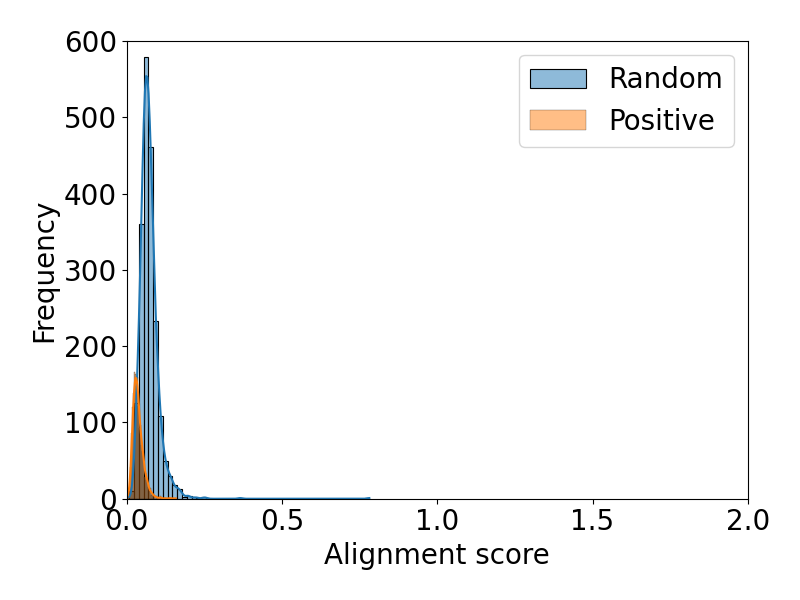}
        \caption*{100k steps in Stage 1}
    \end{minipage}
    \hfill
    \begin{minipage}{0.32\textwidth}
        \centering
        \includegraphics[width=\linewidth]{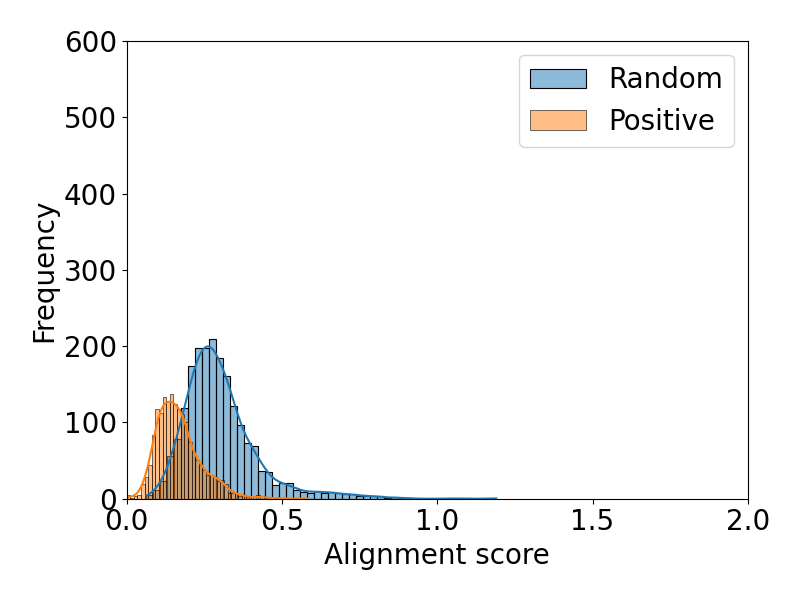}
        \caption*{400k steps in Stage 1}
    \end{minipage}
    \hfill
    \begin{minipage}{0.32\textwidth}
        \centering
        \includegraphics[width=\linewidth]{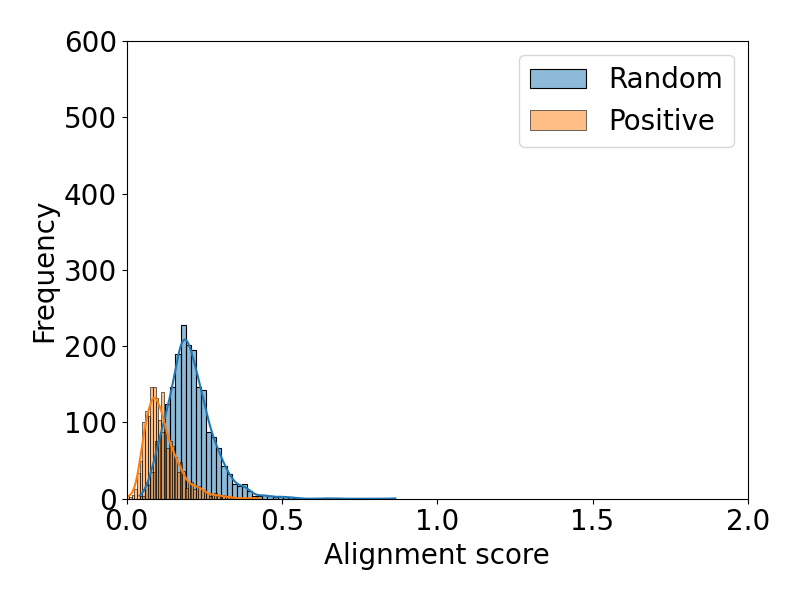}
        \caption*{200k steps in Stage 2}
    \end{minipage}
    \vspace{1em}
    \begin{minipage}{0.32\textwidth}
        \centering
        \includegraphics[width=\linewidth]{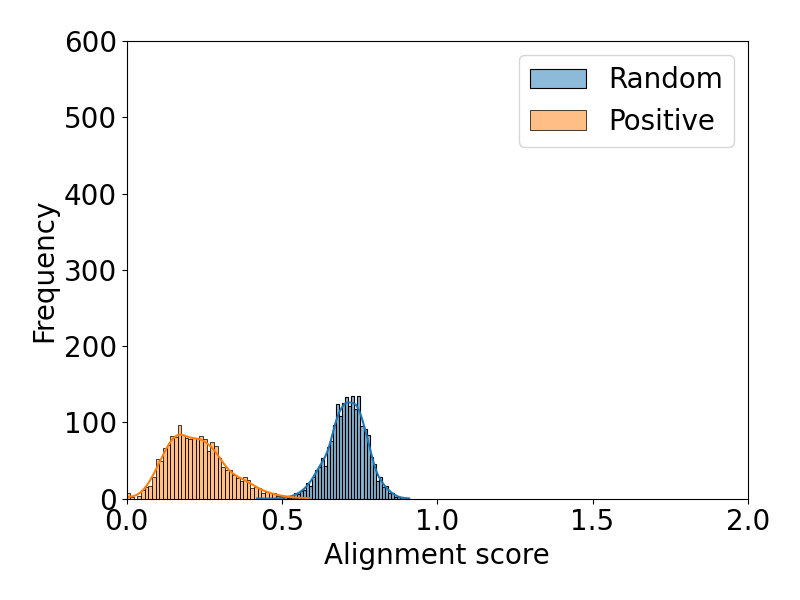}
        \caption*{cl-nagoya/ruri-large-v2}
    \end{minipage}
    \hfill
    \begin{minipage}{0.32\textwidth}
        \centering
        \includegraphics[width=\linewidth]{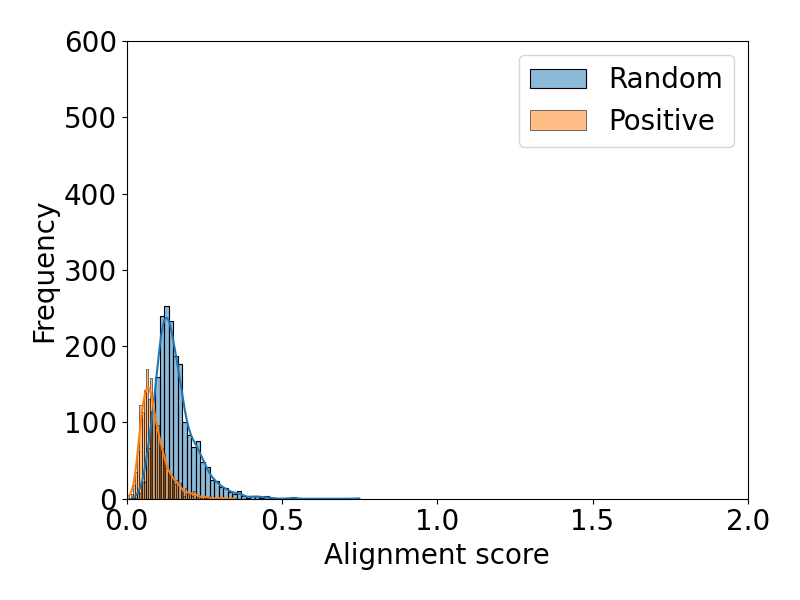}
        \caption*{sbintuitions/modernbert-ja-130m}
    \end{minipage}
    \hfill
    \begin{minipage}{0.32\textwidth}
        \centering
        \includegraphics[width=\linewidth]{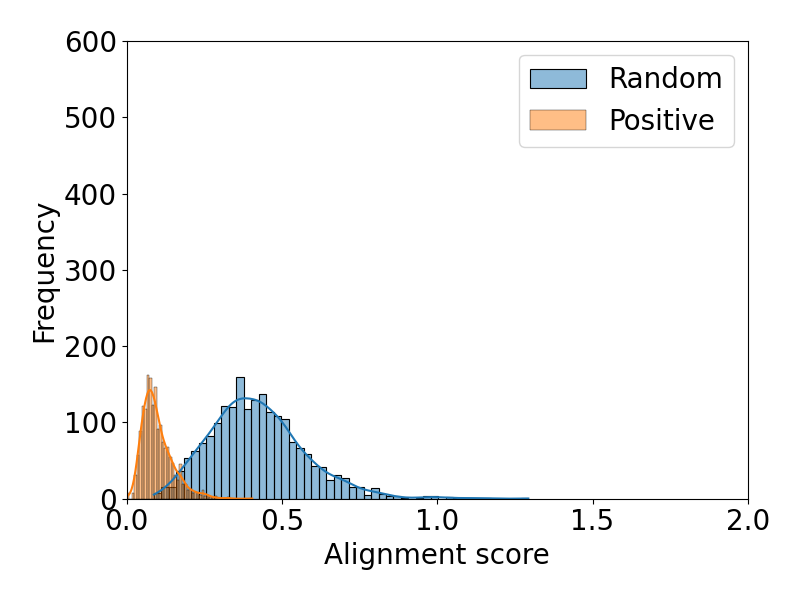}
        \caption*{tohoku-nlp/bert-base-japanese-v3}
    \end{minipage}
    \caption{Distribution of sentence similarities.}
    \label{fig:sim_dist_dynamics}
\end{figure*}

Figure~\ref{fig:alignment_and_uniformity} illustrates the progression of alignment and uniformity at each checkpoint during training. At 0k steps (model initialization), uniformity is low, while alignment is high. This is likely due to the random initialization of parameters, which causes sentence embeddings to be distributed in arbitrary directions. As training progresses through 4k, 15k, and 100k steps, uniformity increases while alignment decreases, suggesting that the embeddings become more biased or anisotropic~\cite{ethayarajh-2019-contextual,gao2019representation,gao-etal-2021-simcse}. Beyond 100k steps, the values fluctuate between those observed at 15k and 100k, reflecting the inherent trade-off between alignment and uniformity.
We also observe that the scores of our model, llm-jp-modernbert, at the final checkpoint in stage 2 (200k steps) are close to those of modernbert-ja-130m, a model that also adopts the ModernBERT architecture.

\subsection{Distribution of Sentence Similarity}
As further analysis, we examine the distribution of cosine similarities for positive and random sentence pairs for each model, using the same dataset as in the alignment and uniformity experiments.

The results are shown in Figure~\ref{fig:sim_dist_dynamics}. Up to 100k steps, the alignment scores the majority of pairs decrease. After that, the distribution of random pairs shifts slightly to the right.
Ruri-large-v2 demonstrates a clear separation between the distributions of positive pairs and random pairs. Similarly, bert-base-japanese-v3 also shows a distinct separation in its distributions. In contrast, modernbert-ja-130m exhibits nearly overlapping distributions for positive and random pairs, similar to the distribution of our model at 200k steps in Stage 2.

\section{Conclusion}

In this paper, we introduced llm-jp-modernbert, a Japanese ModernBERT model trained on a large-scale corpus with a context length of 8192 tokens. While the model does not outperform existing baselines on downstream tasks, it shows good performance on fill-mask test evaluations. 
We also conducted an in-depth analysis using training checkpoints, exploring the impact of context length expansion through pseudo-perplexity and investigating sentence embedding dynamics during training. Our comparisons with existing models show consistent behavior among those with similar architectures.

\section*{Acknowledgments}
We thank Hayato Tsukagoshi, Hiroshi Matsuda, Jiro Nishitoba, and the members of the LLM-jp for their valuable feedback and advice.
In this research work, part of the results were obtained using the ``mdx: a platform for building data-empowered society'' and SAKURA internet Inc.'s ``High Firepower PHY Service''.

\bibliography{custom}

\appendix
\onecolumn

\begin{figure}[t]
    \centering
        \includegraphics[width=0.7\linewidth]{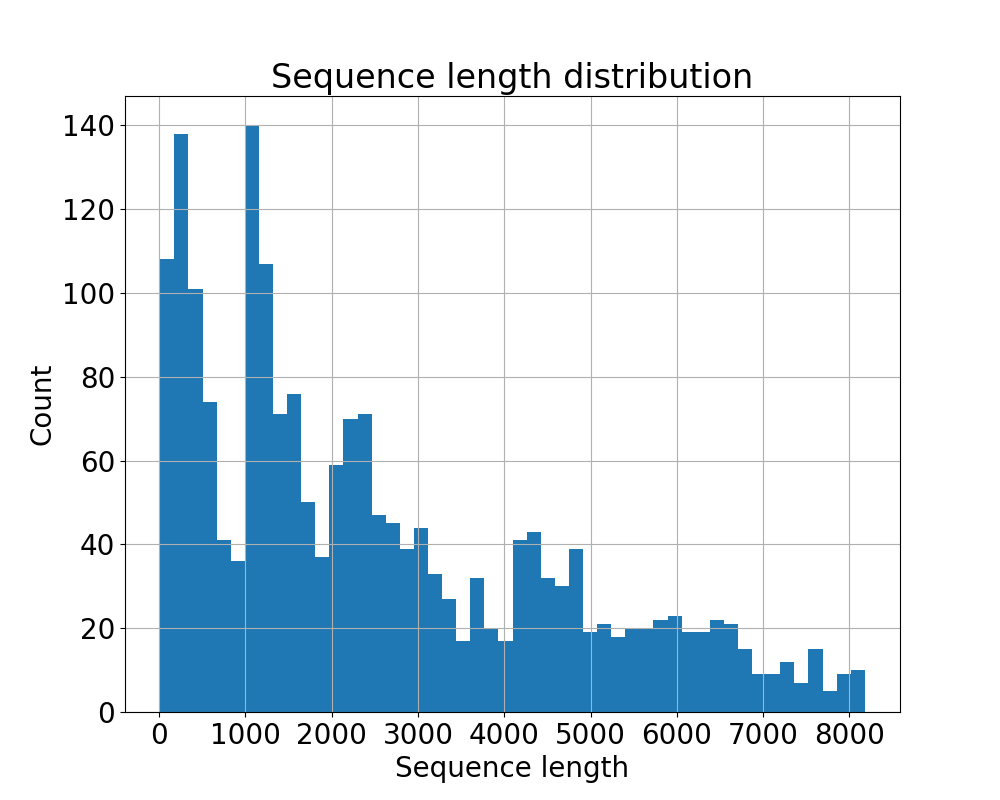}
        \caption{The sequence length distribution of sentences with various sequence lengths ranging from 0 to 8192, prepared for the pseudo-perplexity experiment. The sequence length in this figure refers to the token count obtained using the llm-jp-tokenizer v3.}
        \label{fig:seq_len_dist}
\end{figure}
\section{Distribution of Sequence Lengths}
\label{sec:appendix_pseudo_ppl}
Figure~\ref{fig:seq_len_dist} shows the distribution of sequence lengths in the dataset used in Section~\ref{sec:pseudo_ppl}.

\begin{table}[t]
\centering
\caption{Performance of sentence retrieval on MIRACL.}
\begin{tabular}{lccccc}
\toprule
Model & Recall@10 & MRR@10 \\
\hline
cl-nagoya/ruri-large-v2~\cite{tsukagoshi2024Ruri} & 0.987 & 0.872 \\
tohoku-nlp/bert-base-japanese-v3~\cite{tohoku2023} & 0.740 & 0.529 \\
sbintuitions/modernbert-ja-130m~\cite{modernbert-ja} & 0.506 & 0.334 \\
Edit distance & 0.289 & 0.198  \\
Jaccard distance & 0.031 & 0.021 \\
\bottomrule
\end{tabular}
\label{table:miracl_result}
\end{table}

\section{Details of Sentence Retrieval Task using MIRACL}
\label{sec:appendix_miracl}

We used the Japanese subset of the MIRACL dataset~\cite{zhang2023miracl}. MIRACL is a dataset for multilingual sentence retrieval task. Each instance contains a query, a set of passages related to the query (positive passages), and a set of passages unrelated to the query (negative passages). The Japanese subset consists of 3,477 instances. 
To perform the retrieval task with MIRACL, we prepared the query and corpus using the following method. 
When multiple sentences were present in positive passages, one sentence was added to the query set and another to the corpus set. We then calculated the sentence similarity between all query-corpus pairs and ranked the matching sentences to compute recall and MRR. 

To validate the constructed task for assessing sentence retrieval performance, we evaluate the performance of various approaches, including the supervised fine-tuned sentence embedding model, pre-trained BERT models, and heuristic methods such as edit distance. Table~\ref{table:miracl_result} shows the result. The recall of the cl-nagoya/ruri-large-v2 model approaches a value close to 1.0, while naive methods such as edit distance yield lower values. This indicates that the constructed task is valid for measuring sentence retrieval performance.

\end{document}